\documentclass[final]{cvpr}

\usepackage{times}
\usepackage{epsfig}
\usepackage{graphicx}
\usepackage{amsmath}
\usepackage{amssymb}
\usepackage{xcolor}
\usepackage[linesnumbered,ruled,vlined]{algorithm2e}
\usepackage{tabularx}
\usepackage{booktabs}
\usepackage[misc]{ifsym}

\newcommand{\rpm}{\raisebox{.2ex}{$\scriptstyle\pm$}}

\def\cvprPaperID{21} 
\def\confYear{CVPR 2021}

\usepackage[pagebackref=true,breaklinks=true,colorlinks,bookmarks=false]{hyperref}

\begin{document}

\pagenumbering{gobble}

\title{A Mathematical Analysis of Learning Loss for Active Learning in Regression}

\author{Megh Shukla ${}^{\textrm{\Letter}}$ \hspace*{10mm} Shuaib Ahmed \\
Mercedes-Benz Research and Development India \\
{\tt\small megh.shukla@daimler.com}

}

\maketitle

\begin{abstract}
   Active learning continues to remain significant in the industry since it is data efficient. Not only is it cost effective on a constrained budget, continuous refinement of the model allows for early detection and resolution of failure scenarios during the model development stage. 
   Identifying  and fixing failures with the model is crucial as industrial applications demand that the underlying model performs accurately in all foreseeable use cases. One popular state-of-the-art technique that specializes in continuously refining the model via failure identification is \textit{Learning Loss}\cite{learnloss}. Although simple and elegant, this approach is empirically motivated. Our paper develops a foundation for Learning Loss which enables us to propose a novel modification we call LearningLoss++. We show that gradients are crucial in interpreting how Learning Loss works, with rigorous analysis and comparison of the gradients between Learning Loss and LearningLoss++. We also propose a convolutional architecture that combines features at different scales to predict the loss. We validate LearningLoss++ for regression on the task of human pose estimation (using MPII and LSP datasets), as done in Learning Loss. We show that LearningLoss++ outperforms in identifying scenarios where the model is likely to perform poorly, which on model refinement translates into reliable performance in the open world.\footnote{~\copyright 20xx IEEE. To appear, Proceedings of the 2021 IEEE/CVF Conference on Computer Vision and Pattern Recognition Workshops (CVPRW)}
\end{abstract}

\section{Introduction}

\begin{figure*}[ht!]
    \centering
    \includegraphics[width=0.7\linewidth]{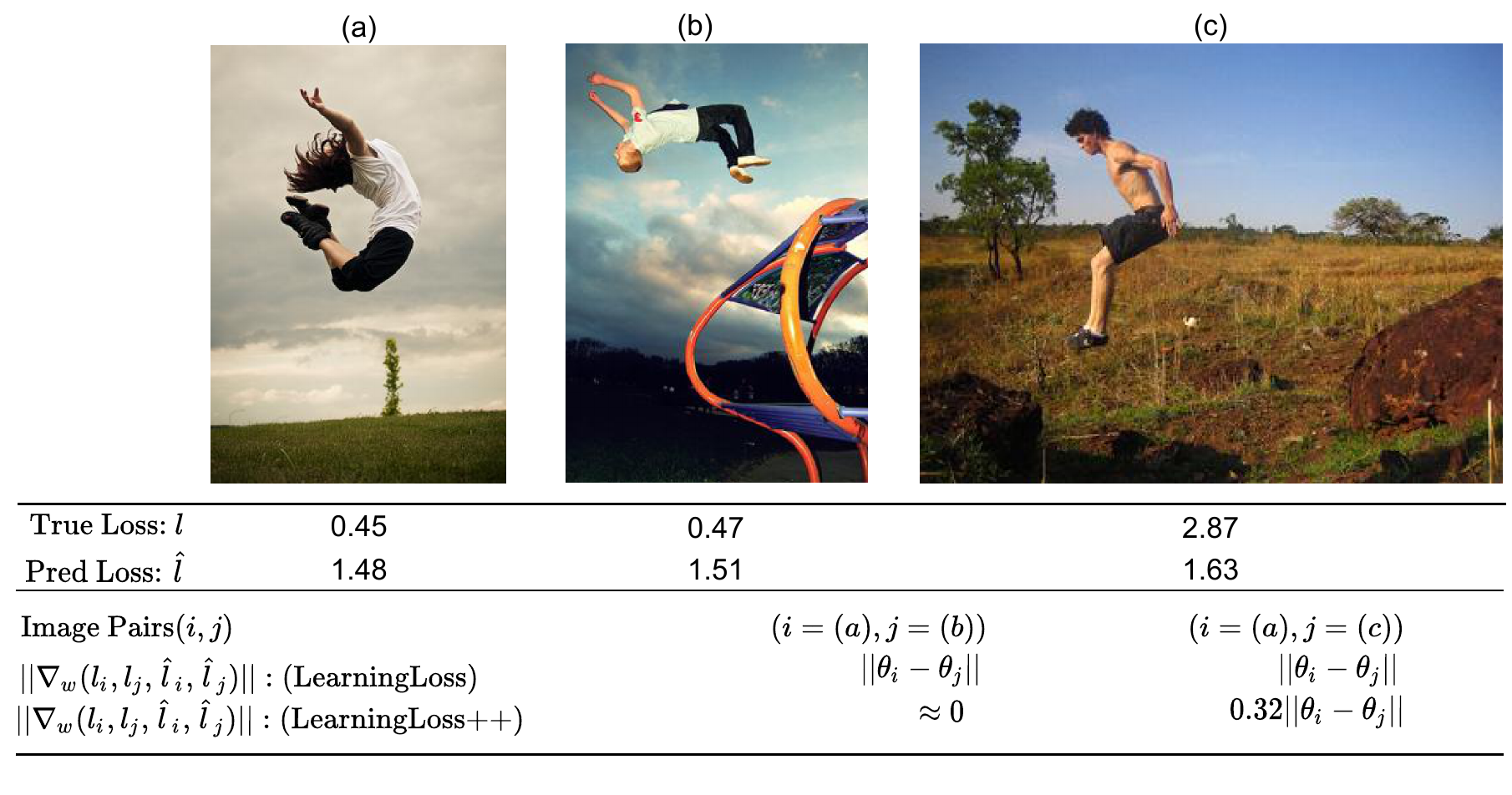}
    \caption{Both Learning Loss and LearningLoss++ (ours) use ranking loss to train the loss predictor network. However, the gradient response for both the approaches significantly differ. For the image pair (a, b), the true loss as well as predicted loss are similar, agreeing with intuition. While LearningLoss++ concurs with this notion, Learning Loss wrongly penalizes the network. For the image pair (a,c), the true loss for (c) is significantly larger than (a) whereas the predicted loss is similar. In this scenario, LearningLoss++ enforces a larger gradient whereas the Learning Loss gradient has the same response as in the previous scenario, in spite of the wrong predictions.}
\label{fig:main}
\end{figure*}

The deep learning era has heralded a paradigm shift in the approaches we take to model our data. Cheaper hardware, clean data and robust modelling has resulted in a large number of industrial applications such as driver monitoring systems and activity recognition. However, like all data-driven paradigms, deep learning has an insatiable appetite for data. Industry applications of human pose estimation often require millions of labelled images for acceptable performance to eliminate risks associated with faulty predictions when the model is deployed on-ground. This introduces a two-fold challenge: finding algorithms to recognize failure cases for the model, as well as reducing the costs associated with the large scale collection of annotated data.

\par
One potential solution lies in active learning, a class of algorithms aimed at reducing annotation costs involved in training the model. The essence behind all active learning algorithms is to allow the model to choose a set of images which would impart maximum information about the dataset if labelled. Although active learning is well explored, active learning in human pose estimation remains a challenge. This is because human pose models \cite{hg, hrnet, mrf} regress two dimensional heatmaps to estimate the location of joints, making this approach different from those tasks having a probabilistic interpretation of outputs. The task of regressing heatmaps is sufficiently different from regressing the joint coordinates \cite{handpose}; with DeepPose (2014) \cite{deeppose} being the last significant technique to do so in human pose estimation.

\par
This paper builds upon Learning Loss For Active Learning \cite{learnloss}, a task agnostic active learning algorithm having performance comparable with state-of-the-art in classification, object detection and human pose estimation. This is remarkable, as we do not specify domain constraints such as geometry of the human skeletal structure explicitly. However, this approach has two drawbacks: 1) The learning loss objective is intuitively defined and 2) The architecture of the learning loss module discards spatial information. With LearningLoss++, our contributions include:
\begin{enumerate}
     
    \item We establish equivalency between Learning Loss' empirically driven objective and the KL divergence objective proposed by LearningLoss++ which is hyperparameter free. 
    
    \item We analyze the gradient to provide intuition into the training process of Learning Loss. We then compare the expected gradients for Learning Loss and LearningLoss++, which allows us to prove the advantages associated with smoothened gradients. 
    
    \item A convolutional architecture to combine features at different scales, replacing the global average pooling - fully connected architecture defined in Learning Loss.
\end{enumerate}

We perform experimental validation using the MPII \cite{mpii} and LSP-LSPET \cite{lspet} datasets and compare our approaches with other methods, with an emphasis on the ability of LearningLoss++ to identify failures. Successful identification of failure cases translates into continuous refinement of the model and thus improved reliability in the open world use cases.

\section{Related Work}

The early foundation for active learning is summarized in \cite{settles2009active}, with approaches involving uncertainty studied in \cite{ent2, uncertainty_segmentation, ent1, gan_multiclass}. Query Based Committee \cite{qbc} and other ensemble approaches \cite{beluch2018power, ensemble} explore the performance of a group of models with different parameter spaces for an image and compute the level of agreement between the models for the image. Expected Gradient Length based approaches \cite{egl1}, \cite{egl2} use gradient as a measure for determining the \textit{informativeness} of an image. Diversity promoting approaches such as \cite{div1}, \cite{div2} and \cite{sener2017active} use a small sample set to represent the larger pool of unlabelled data for annotation. Bayesian active learning approaches \cite{gal2017deep}, \cite{kendall2017uncertainties} provide beautiful theoretical results using aleatoric and epistemic uncertainties, but to the best of our knowledge, there are no available results for human pose estimation.

\par
Bayesian uncertainty has been used in hand pose estimation (DeepPrior) \cite{handpose} but a direct application is not possible for human pose estimation. DeepPrior directly regresses the joint coordinates, unlike human pose estimation which regresses entire heatmaps. Estimating the epistemic uncertainty for entire heatmaps is not studied, limiting its application to human pose estimation. Since DeepPrior consists of fully connected layers, applying dropout is a standard technique. However, the use of dropout is absent in fully convolutional architectures such as those used in human pose estimation. Finally, the extension of DeepPrior when joints are occluded is not clear. Aleatoric uncertainty has been used in human pose estimation \cite{ajain} to estimate the location of occluded joints, however this approach does not extend to active learning.

\par
Extending these active learning approaches to human pose estimation is not trivial. Traditional uncertainty based approaches find applications in classification due to the availability of a posterior distribution, which is unavailable for human pose estimation. Ensemble approaches are memory intensive for deployment on edge devices. Methods relying on gradient length are computationally expensive which limit their use in real-time applications. Diversity techniques lack the ability to detect use cases where the model is likely to fail. Bayesian uncertainty approaches not only restrict the network to a Bayesian Convolutional Network (BCN), they  rely on the use of Dropout and multiple forward propagation runs reducing their use in real-time applications. The first algorithms dedicated to active learning for human pose estimation was \cite{liu2017active} which involved the use of multi-peak entropy. It can be shown that multi-peak entropy value is essentially a lower bound for the standard entropy approach. Learning loss for active learning \cite{learnloss} proposed the use of a general purpose auxiliary model trained to predict an indicative loss for tasks involving classification, detection and pose estimation. The underlying philosophy for learning loss is to detect images where the model performs poorly. Our work builds upon this paper to improve the correlation between predicted and true loss which enables better detection of failures therefore improving the model early on during the development phase.

\section{Learning Loss (Yoo and Kweon) \cite{learnloss}}

The learning loss module is an auxiliary network $h(.)$ attached to the intermediate layers of the main model (classifier/pose estimator) $f(.)$. For a given image $x$, we have: $\hat{y} = f(x)$ where $\hat{y}$ is the prediction of the model $f$ with the ground truth $y$. The  value of true loss $l$ is $l = L_{model}(y, \hat{y})$ where $L_{model}(.)$ is the loss function such as cross-entropy / mean square error. Intermediate representations of the model $f^{h}(x)$ are inputs for the learning loss network to predict a '\textit{loss}' $\hat{l}$, where $\hat{l} = h(f^{h}(x))$.

\subsection{Method}

The notion of predicted loss $\hat{l}$ is similar to that of the true loss $l$, with a high value of $\hat{l}$ implying that the model has likely produced a wrong prediction. In the absence of true loss as is the case with unlabelled images, the predicted loss steps in to compute the performance of the model on the input image. Therefore, images with a high $\hat{l}$ are selected for annotation to refine the use cases where the model is predicted to perform poorly.  To train the learning loss network, the authors use a ranking loss to compare a pair of images. Let $(x_i, l_i, \hat{l}_i)$, $(x_j, l_j, \hat{l}_j)$ represent a pair of (image, true loss, predicted loss). The objective to train the learning loss network is:
\begin{equation}
    \mathbb{L}_{loss} = \textrm{max}\,\Big( 0, -\,\textrm{sign}(l_i - l_j)(\hat{l}_i - \hat{l}_j) + \xi \Big)
\label{eq:og}
\end{equation}
The idea behind Eq: \ref{eq:og} is simple, if the true loss $l_i$ corresponding to $x_i$ is greater than $l_j$, then the predicted loss $\hat{l}_i$ has to be greater than $\hat{l}_j$ by a margin $\xi$ so as to not incur any loss. Similarly, if $l_j > l_i$ then $\hat{l}_j > \hat{l}_i$. The learning loss network incurs a loss if the predicted loss is not greater than $\xi$ or if the learning loss network predicts the opposite ($l_i > l_j$ but $\hat{l}_i < \hat{l}_j$). The decision to use a pairwise comparison of images might seem strange, especially when mean square error $(l - \hat{l})^2$ can be used to train the learning loss network to learn the mapping between an input image $x$ and the true loss $l$. The authors on the contrary argue that the network trained on MSE fails to learn anything meaningful in their experimental study.
\par
Active learning using learning loss is straighforward. We train the task specific model $f(.)$ and the learning loss network $h(.)$ using all labelled images $(x \in \mathcal{L})$. We then select a subset of unlabelled images ($x \in \mathcal{U}$) for annotation which have a high predicted loss $\hat{l} = h(f^{h}(x))$. This process continuous cyclically for continuous refinement of the model.

\section{LearningLoss++}

We first establish equivalency between the intuition driven objective (Eq: \ref{eq:og}) and the KL divergence based objective of LearningLoss++ by comparing the gradients. We then analyse the gradient formulation to provide insights into the training of the learning loss network. This is followed by highlighting some shortcomings associated with the Learning Loss gradient, and finally we analyse the LearningLoss++ gradient to show that it implicitly absorbs the margin $\xi$ hyperparameter.
\subsection{Gradient Analysis}
\begin{figure}
        \begin{minipage}{1.0\linewidth}
        \centering
        \centerline{\includegraphics[width=6cm]{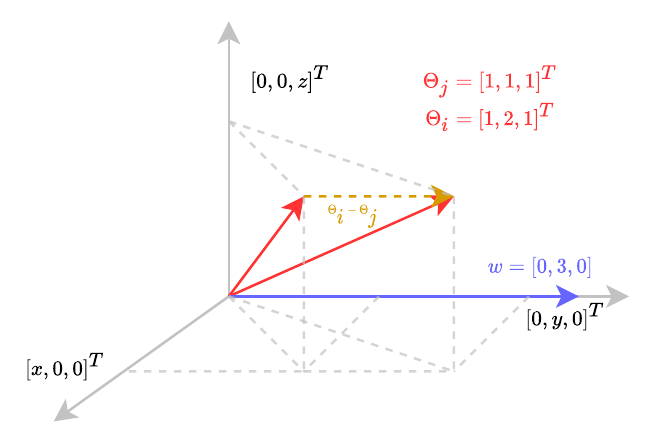}}
        \end{minipage}
        \caption{We use an example to demonstrate that Learning Loss forces the weight vector $w$ to align along the most discriminative component $\theta_i - \theta_j$ to explain away the predicted losses}
        \label{fig:vector}
\end{figure}
Let $\theta$ represent the output of the \textit{penultimate layer} in the fully connected learning loss network. By definition, the predicted loss in terms of $\theta$ is: $\hat{l} = \theta^Tw$ where $w$ represents the weights of the final learning loss layer. While we consider the gradients associated with the penultimate layer, we note that these gradients are backpropagated and hence our analysis extends to any general fully connected network.
\par
\textbf{Gradient for Learning Loss}: The learning loss objective Eq: \ref{eq:og} can now be written as:
$\mathbb{L}_{loss} = \textrm{max}\,( 0, -\,\textrm{sign}(l_i - l_j)(\theta_i^T w - \theta_j^T w) + \xi )$. The gradient for learning loss is:
\begin{subequations}
    \begin{align}
        \nabla_w \mathbb{L}_{loss}  & \in \{0,  \rpm (\theta_i - \theta_j)\} \\
        \nabla_{\theta} \mathbb{L}_{loss}  & \in \{0, \rpm w\}     
    \end{align}
\label{eq:og_grad}
\end{subequations}

\vspace*{-2mm}
Eq: \ref{eq:og_grad} provides an interesting insight into the behaviour of the module. \textit{The weights Eq: \ref{eq:og_grad}(a) are aligned to emphasize on the most discriminative component between the intermediate features}. If we treat $\theta$ as a vector in $n-dimensional$ space and $\theta_i$ and $\theta_j$ differ along one component only, the weight vector amplifies the response along that particular component to predict the indicative loss, as we discuss below.
\par
\textbf{So How Does Learning Loss work?} We highlight the discriminative property of the gradients with an example. For this example, let us assume $\theta_i = [1, 2, 1]^T$, $\theta_j = [1, 1, 1]^T$ and $w = [0, 3, 0]$ in a three dimensional vector space as shown in Fig: \ref{fig:vector}. We use the same value of margin $\xi$ as recommended in the paper. Depending upon the true loss ($l_i$ and $l_j$), the objective leads us to two cases: 1) $l_i > l_j$ and 2) $l_i < l_j$. Substituting the values of $\theta_i, \theta_j$ and $w$ in $\mathbb{L}_{loss}$ (Eq: \ref{eq:og}) for case 1, we get: $\mathbb{L}_{loss}(w, \theta_i, \theta_j) = max\,(0, -1*(6 - 3) + 1)$ = 0.
\par
Intuitively, this corresponds to taking a projection of both the penultimate layer outputs $\theta$ along the weight vector $w$. Since the current weight vector is already aligned along the discriminative component, the learning loss model does not incur a penalty. Things get interesting if we consider the second case, $l_i < l_j$. In this scenario, the objective incurs a penalty ($\mathbb{L}_{loss} = max\,(0, 1*(6 - 3) + 1)$ = 4). Using Eq: \ref{eq:og_grad}(a), we get $\nabla_w \mathbb{L}_{loss} = [0, 1, 0]^T$, with gradient descent acting against this direction ($-\nabla_w \mathbb{L}_{loss}$). This step enables the weight vector to reverse direction and ultimately point along the direction that minimizes the loss. \\
\par
\textbf{Gradient for LearningLoss++}: As we saw in the previous example, the discriminative property $(\theta_i - \theta_j)$ of the gradient allows the learning loss network to align weights in a manner to explain the predicted losses. We show that the proposed KL divergence based training objective has a gradient with a similar form.
\par
As before, let $\hat{l}_i = \theta_i^Tw$ and $\hat{l}_j = \theta_j^Tw$ represent the predicted loss for images $i$ and $j$. We now consider two images in the minibatch and compute softmax over $\hat{l}$: $q_i = e^{\hat{l}_i} / (e^{\hat{l}_i} + e^{\hat{l}_j})$ with $q_j$ defined similarly. This interpretation can be viewed as the probability of sampling $x_i$ over $x_j$ and vice versa for annotation. While we have defined a probabilistic interpretation for the predicted losses, a similar one needs to be defined for the true losses. Since true losses ($l_i, l_j > 0$), we use simple scaling $p_i = \dfrac{l_i}{l_i+l_j}$ to denote the probability of $x_i$ having a higher true loss than $x_j$. Intuitively, an image $x_i$ having a true loss $n$-times greater than $x_2$ is n-times more likely to get sampled for annotation. The objective to minimize is:
\begin{equation}
    \mathbb{L}_{loss}(w, \theta_i, \theta_j) = \textrm{KL}(p||q) = p_i \textrm{log}\dfrac{p_i}{q_i} + p_j \textrm{log}\dfrac{p_j}{q_j}
    \label{eq:plus_1}
\end{equation}
For brevity, we reproduce the final solution, referring the reader to the supplementary material for the full derivation. The solution is simple and delightfully familiar:
\begin{subequations}
    \begin{align}
        \nabla_w \mathbb{L}_{loss}(w, \theta_i, \theta_j) & = (q_i - p_i)(\theta_i - \theta_j) \\
        \nabla_\theta \mathbb{L}_{loss}(w, \theta_i, \theta_j) & = (q_i - p_i)w     
    \end{align}
\label{eq:plus_grad}
\end{subequations}

\vspace*{-5mm}
\textbf{Active Learning with LearningLoss++}: Although we have introduced softmax and KL divergence with LearningLoss++, the process of active learning sampling remains the same as in Learning Loss. The images corresponding to the top-k predicted losses $\hat{l}$ are chosen for annotation. We use softmax and  KL divergence to train the loss prediction network only. Therefore, while the gradient computation is probabilistic, the core active learning sampling process is deterministic.

\subsubsection{LearningLoss++ Advantages}

\begin{figure}
        \begin{minipage}{1.0\linewidth}
        \centering
        \centerline{\includegraphics[width=8cm]{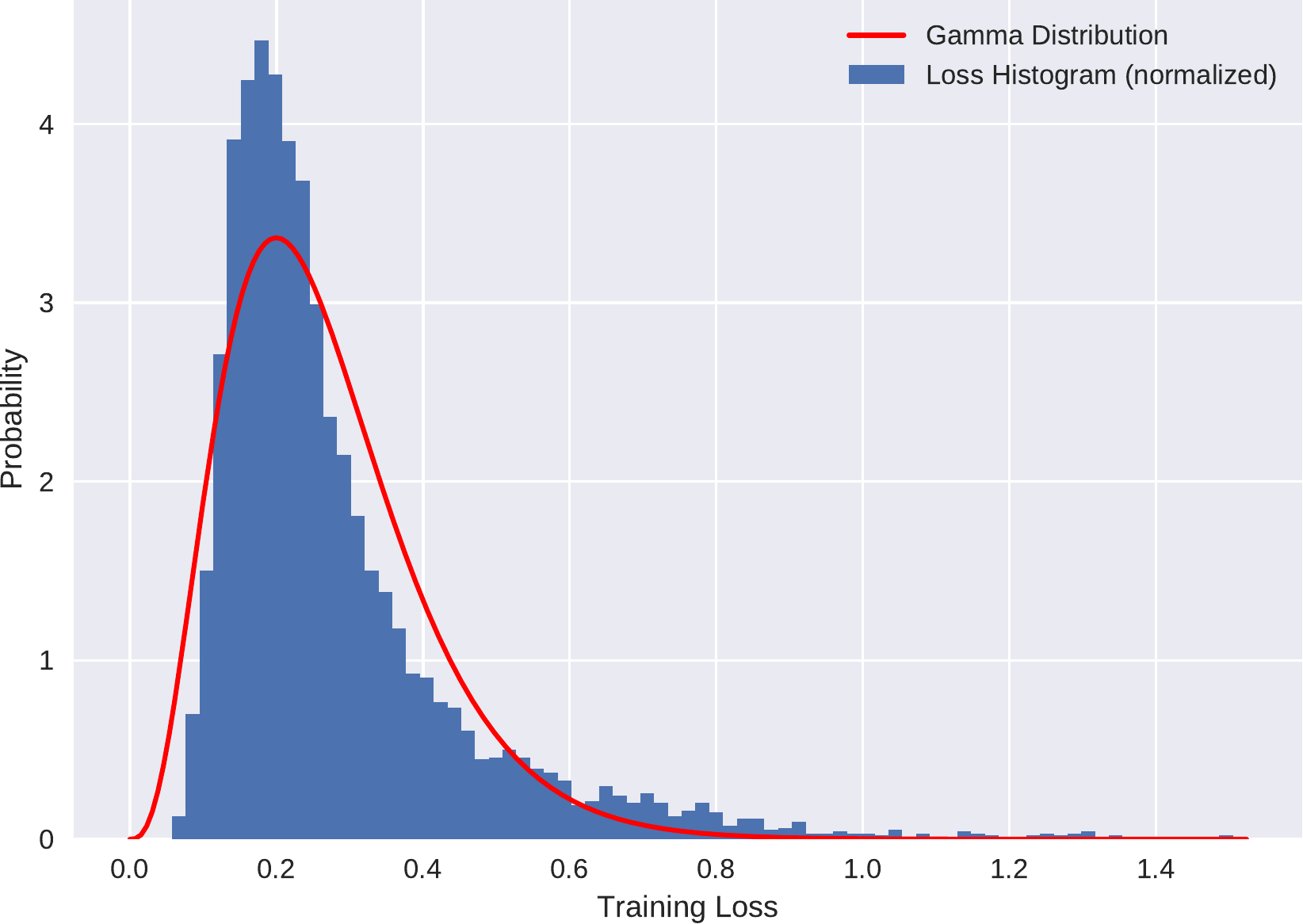}}
        \end{minipage}
        \caption{Statistical results show that the training losses (true loss) follow a gamma distribution for a sufficiently converged regression model. This figure shows the true loss histogram for a human pose estimation model. We can approximate this particular histogram with $\gamma(k=4, \Theta=0.066)$ obtained using maximum likelihood estimation.}
        \label{fig:loss_dist}
\end{figure}

We have succesfully proven that Learning Loss and LearningLoss++ share the same underlying principles in training. The most prominent feature of the LearningLoss++ gradient is the presence of $(q_i - p_i)$ that smoothens the gradient in comparison to Learning Loss. This smoothening, as we will show, allows for better detection of \textit{lossy} images. We additionally note that LearningLoss++ implictly absorbs the margin hyperparameter $\xi$.\textbf{ Fig: \ref{fig:main} depicts the two scenarios which highlight the advantages associated with the LearningLoss++ gradient}. We discuss the two scenarios in detail: \\
\par
\textbf{Scenario 1} (Fig: \ref{fig:main}, image pair-(a, b)): In the event we sample a pair of images with similar true losses, Learning Loss incurs a loss even when the predicted losses are similar, since the margin $\xi$ forces the predicted losses to have a prediction loss margin of at least $\xi$. With LearningLoss++, the network does not incur a penalty. Since $l_i \approx l_j \implies p_i \approx p_j \approx 0.5$ and $\hat{l}_i \approx \hat{l}_j \implies q_i \approx q_j \approx 0.5$, we have $(q_i - p_i) \approx 0$. Although one question remains; how likely are we to sample a pair of images having similar true losses? Or formally, if $P(X=l_i, Y=l_j)$ represents the probability of sampling images with losses $(l_i, l_j)$, then given a true loss margin $\delta$ (not to be confused with $\xi$), what is $P(|X-Y|\le \delta)$?

Fortunately, a closed form solution exists to compute the probability of sampling a pair of images with similar losses. Statistical regression under the condition of homoscedasticity assumes that the residuals (losses) are distributed as per the normal distribution $\varepsilon \sim \mathcal{N}(0, \sigma^2)$. The distribution of squared residual $\varepsilon^2$ follows $\mathcal{N}^2(0, \sigma^2) = \gamma(\frac{1}{2}, 2 \sigma^2)$. This result can be generalized to a summation of $n$ gaussian distributions: $p(\varepsilon^2) = \sum_{i=1}^{n}\mathcal{N}_i(0, \sigma^2) = \gamma(\frac{n}{2}, 2 \sigma^2)$. Therefore, the probability of sampling a loss follows a gamma distribution. : $P(X=l_i) = \gamma(k, \Theta)$ as shown in Fig: \ref{fig:loss_dist}.
\par
The probability of sampling two images with true losses within $\delta$ is:
\begin{align}
        P(|X - Y| \le \delta) & = \int_0^{\delta}\gamma(x, k, \Theta)\int_0^{x+\delta}\gamma(y, k, \Theta)\mathrm{d}y\mathrm{d}x \nonumber \\ &+ \int_{\delta}^{\infty}\gamma(x, k, \Theta)\int_{x-\delta}^{x+\delta}\gamma(y, k, \Theta)\mathrm{d}y\mathrm{d}x
    \label{eq:xy_le_del}
\end{align}%

To compute a closed form solution for Eq: \ref{eq:xy_le_del}, we restrict $k \in \mathbb{Z}^{+}$ allowing us to compute the integral of the gamma distribution analytically. Since the rest of the derivation involves extensive simplification, we request the reader to refer to the final result Eq: (5) from the supplementary material. To provide intuition into Eq:5-supplementary, we use the loss distribution from Fig: \ref{fig:loss_dist} as an example to compute the $P(|X-Y| \le \delta)$ for various true loss margin $\delta$ in Table: \ref{tab:closed_loss}. We observe that there is a high probability (10\% - 43\%) of sampling a pair of images having similar true losses. For a well trained loss prediction module that correctly predicts similar $\hat{l}$ for similar $l$, this results in upto 10\% - 43\% of the resultant gradient updates being noisy. Fortunately, LearningLoss++ does not suffer from this issue.\\

\begin{table}
\centering
\resizebox{\columnwidth}{!}{%
\begin{tabular}{lrrrrrrr}
\toprule
$\delta$  & 0.02 & 0.04 & 0.06 & 0.08 & 0.1 & 0.125 & 0.15\\
\midrule
$P_{X, Y, \gamma}$  & 0.094  & 0.185 & 0.274 & 0.358 & 0.437 & 0.527 & 0.607\\

\bottomrule
\end{tabular}
}
\vspace*{2mm}

\caption{We compute $P(|(X=l_i) - (Y=l_j)| \le \delta)$ using the closed form solution (Eq:5-supplementary) for true loss - $l_i, l_j$ distributed according to $\gamma(k=4, \theta=0.066)$. We have verified the correctness of our solution with a computer simulation.}
\label{tab:closed_loss}

\vspace*{-2mm}
\end{table}

\begin{figure*}[ht!]
    \centering
    \includegraphics[width=0.7\linewidth]{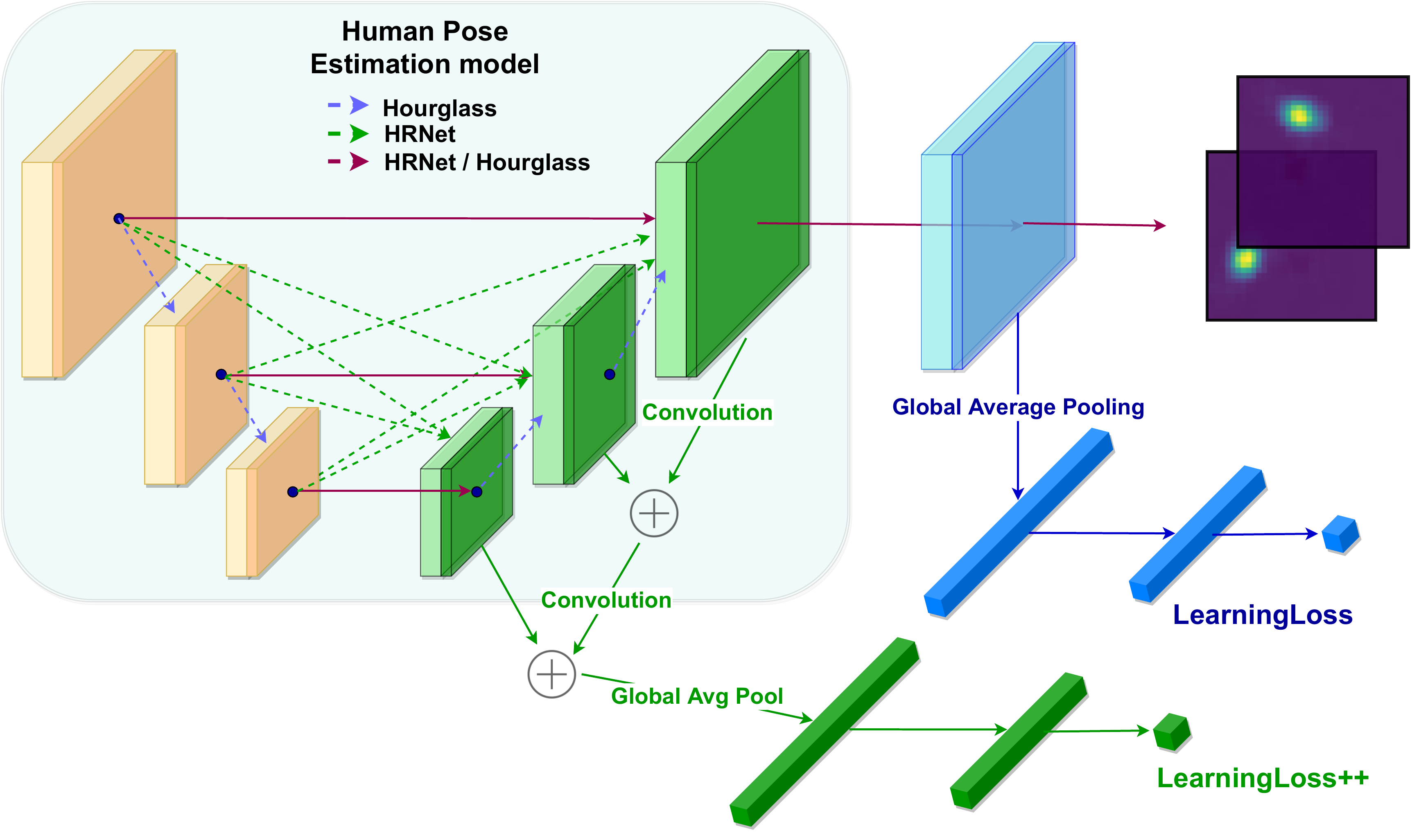}
    \caption{Learning Loss uses global average pooling to collapse the spatial dimensions of the intermediate feature maps into a vector. We argue that such an approach helps only in trivial cases: when the losses can be inferred from cues such as image backgrounds. Human pose estimation relies on spatial interaction between various features, which is lost with pooling. A convolutional feature extractor for learning loss hence captures features at multiple scales without spatial loss, allowing the learning loss network to perform well even when analyzing complex poses.}
\label{fig:arch}
\end{figure*} 

\par
\textbf{Scenario 2} (Fig: \ref{fig:main}, image pair-(a, c)): We also consider the case where the true losses $(l_i, l_j)$ for a pair of images are significantly different and their predicted losses $(\hat{l}_i, \hat{l}_j)$ are similar. In this scenario, Learning Loss correctly incurs a gradient, however the gradient formulation is the same as when the true and predicted losses were similar. \textit{LearningLoss++ gives greater weightage to the gradient when the predicted losses for a pair of images do not reflect the fact that one of the images sampled in the pair has a high true loss than the other}. To prove this statement, we compute the expected gradient (Eq: \ref{eq:plus_grad}) for LearningLoss++ for a fixed true loss margin $\delta$. For a true loss pair $(X=l_i, Y=l_j)$ sampled from $\gamma (k, \Theta)$ where $Y = X + \delta$ (results hold true when $Y = X - \delta$ by symmetry) and $p_i = l_i/(l_i+l_j)$, the expected gradient with respect to $\gamma_{k, \theta}$ is:

\begin{align}
    & \mathbb{E}_{x, y | \delta_2} \Big[ \nabla_w \mathbb{L}(w, \theta_i, \theta_j) \Big] = \lim_{\delta_1 \to \delta_2} \int_{x=0}^{x= \infty}\int_{y=x+\delta_1}^{y=x+\delta_2} \nonumber \\
    & (q_i - \dfrac{x}{2x + \delta_2})(\theta_i - \theta_j)\dfrac{\gamma(x,k,\Theta) \gamma(y,k,\Theta)}{p(y-x=\delta_2)} \mathrm{d}y \mathrm{d}x
    \label{eq:expected_1}
\end{align}
The continuous nature of the gamma distribution leads us to use $\delta_1 \rightarrow \delta_2$ to faithfully infer area under the curve as probability. The simplification of Eq: \ref{eq:expected_1} is along similar lines as deriving $P(|X-Y| \le \delta)$, hence we refer the curious reader to the supplementary material for a complete derivation. We note that the final solution (Eq: 12-supplementary) can be written as a function of true loss margin $\delta$: $\mathbb{E}_{x,y|\delta}[\nabla_w \mathcal{L}] = (q_i - \phi(\delta))(\theta_i - \theta_j)$. Since Eq: 12-supplementary is verbose and not very intuitive, we turn to Table: \ref{tab:grad} for a more intuitive outlook. When $\delta = 0$, the loss prediction network incurs no loss if it predicts $\hat{l}_i \approx \hat{l}_j \implies q_i \approx q_j \approx 0.5$. However, a response of $q_i \approx q_j \approx 0.5$ when the true loss margin $\delta = 0.5$ incurs a large larger gradient response: $(0.5 - 0.18)(\theta_i - \theta_j)$. Only when the loss predictor network predicts $\hat{l}_i < \hat{l}_j \implies q_i < q_j$, the network incurs a lower gradient response. Therefore, for a pair of images where one image has a higher true loss, the loss predictor network is forced to predict a high value of predicted loss $\hat{l}$ for that particular image, in the absence of which the gradient penalty is steep. This translates into the network better identifying images with a high true loss (or faulty inferences). We also note that the softmax $(q = \textrm{softmax}(\hat{l}_i, \hat{l}_j))$ associated with LearningLoss++ removes the need for a predicted loss margin $\xi$ hyperparameter. 
\par

\begin{table}
\centering
\resizebox{\columnwidth}{!}{%
\begin{tabular}{lrrrrrr}
\toprule
$\delta \rightarrow$  & 0.0 & 0.1 & 0.2 & 0.3 & 0.4 & 0.5 \\
\midrule
LL++  & $q_i \textrm{-} 0.5$  & $q_i \textrm{-} 0.39$ & $q_i \textrm{-} 0.3$ & $q_i \textrm{-} 0.25$ & $q_i \textrm{-} 0.21$ & $q_i \textrm{-} 0.18$ \\
LL  & \multicolumn{6}{c}{$\leftarrow \textrm{constant }c_1 \rightarrow$} \\
\bottomrule
\end{tabular}
}
\vspace*{2mm}
\caption{We show the expected gradient response $K(\theta_i - \theta_j)$ where K is tabulated above for different values of the true loss margin $\delta$ using Eq:12-supplementary. These values are computed assuming that the true loss values $l$ are distributed according to $\gamma(k=4, \Theta=0.066)$ (Fig: \ref{fig:loss_dist}). LL++: LearningLoss++, LL: Learning Loss}
\label{tab:grad}
\vspace*{-2mm}
\end{table}

\par
So far, we have rigorously analyzed learning loss for regression. We have successfully shown that the objectives of Learning Loss and LearningLoss++ are equivalent. We have highlighted the role played by the gradient in aligning the weights in a manner that explains the predicted loss. We have shown that a non-trivial number of Learning Loss gradient updates are noisy. We later derived the expected gradient formulation for LearningLoss++, which ensures that the images with a high true loss are identified correctly. Our discussion till now on Learning Loss and LearningLoss++ was valid for all regression tasks. In the next section, we propose a convolutional architecture for the loss prediction module and discuss its associated advantages specific to human pose estimation. 
\begin{figure*}[ht!]
    \hspace*{4em}
    \includegraphics[width=0.735\linewidth]{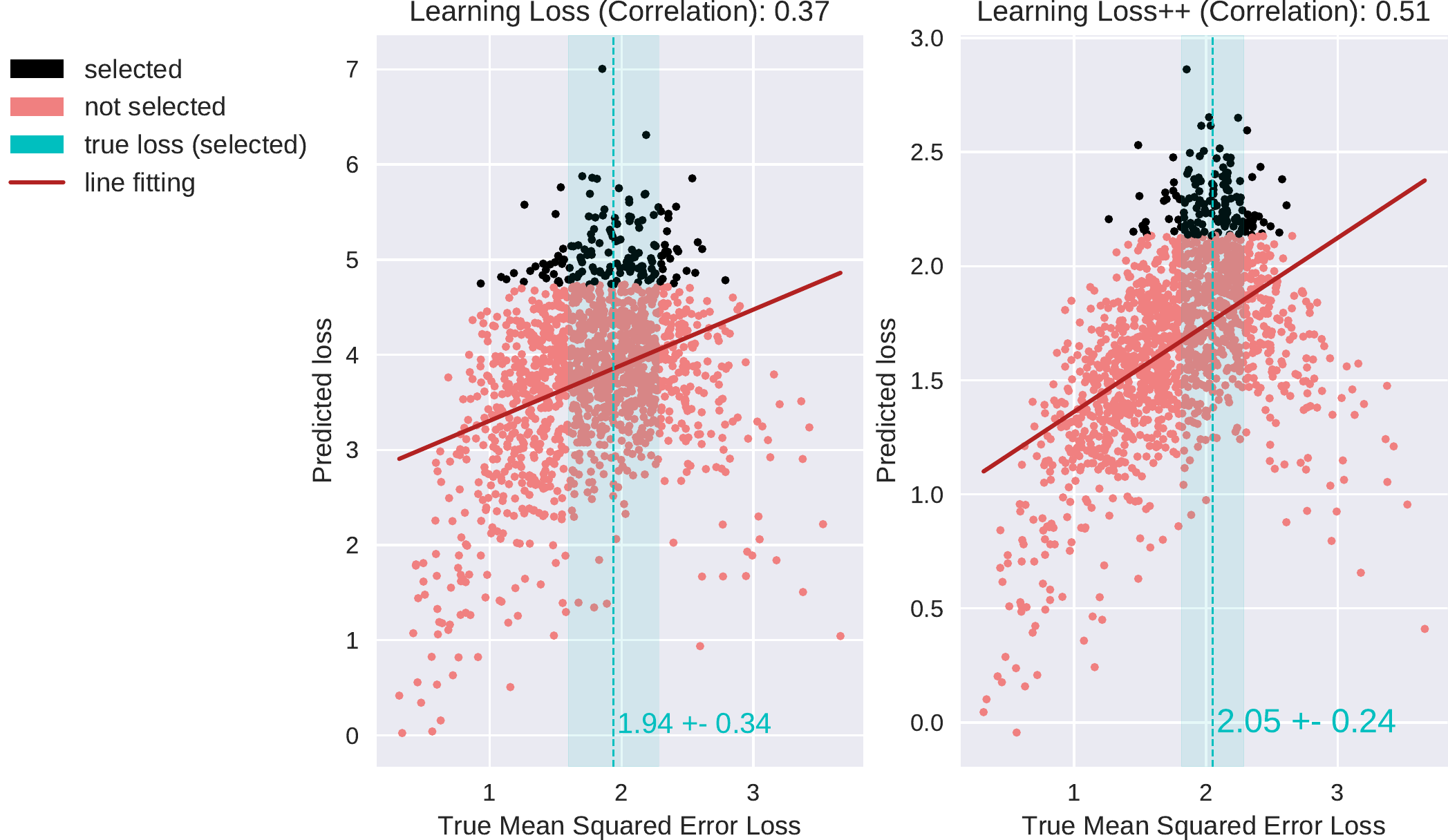}
    \caption{LearningLoss++ yields a higher degree of correlation with the true loss. Not only is the mean loss for the sampled images (black) higher, the true loss variance within the sampled points is lower. (Best viewed when zoomed)}
\label{fig:corr}
\end{figure*}
\subsection{Convolutional Architecture: \\ \hspace*{-3mm} Human Pose Estimation}

Instead of using global average pooling (GAP) (Fig: \ref{fig:arch}), we suggest the use of a convolutional feature extractor for LearningLoss++, since GAP removes any spatial dependency between the features which the learning loss model can use to predict the indicative loss. The convolutional network takes as input intermediate features at multiple scales from the human pose estimation model. These multi-resolution intermediate representations are a vital feature of state-of-the-art models such as HRNet and Stacked Hourglass that allow the network to have a microscopic as well as a macroscopic view of the images. 
\par
We represent the multiscale representations of spatial dimensions $n \times n$ as $\mathcal{H}_{n \times n}$. The stacked hourglass has features at five such scales in an hourglass: $\mathcal{H}_{4 \times 4}, \mathcal{H}_{8 \times 8} \dots \mathcal{H}_{64 \times 64}$. To combine features at multiple scales, we suggest the use of strided convolutions. Let $\mathcal{H}_{n1 \times n1}, \mathcal{H}_{n2 \times n2}$ represent \textit{features at consecutive scales} where $n1 > n2$. Then, the combination of the two features is $\mathcal{H}_{n2 \times n2} \mathrel{+}= \textrm{Conv}( \textrm{stride} = \frac{n1}{n2}, \textrm{kernel\_size} = \frac{n1}{n2}) (\mathcal{H}_{n1 \times n1})$. If $n1$ is not perfectly divisible by $n2$, the convolution takes place with stride, kernel =$ \lfloor \frac{n1}{n2} \rfloor$ followed by another convolution with stride=1 and an appropriate kernel size chosen so that the output of this convolution matches $\mathcal{H}_{n2 \times n2}.$ \textit{We perform a superimposition $\mathcal{H}_{n2 \times n2} + (\mathcal{H}_{n1 \times n1} \xrightarrow[]{Conv} \mathcal{H}_{n2 \times n2})$ as concatenation would drastically increase the size of the network.} The reduction process from $\mathcal{H}_{n1 \times n1} \rightarrow \mathcal{H}_{n2 \times n2} \dots$ is carried out till we reach the final representation (smallest spatial dimension) after which we use a global average pooling to reduce the features to a one dimensional vector. This is followed by a standard fully connected network to infer the predicted losses.

\section{Experimentation and Result}

Our code (written in PyTorch \cite{PyTorch}) will be made available on \url{https://github.com/meghshukla/math-analysis-learningloss} and we use open sourced code wherever possible. We report results on the LSP-LSPET datasets and MPII datasets (as done in \cite{liu2017active, learnloss}) with the following experiments: 1) Correlation between predicted loss and true loss \cite{learnloss} 2) Simulating 5-7 active learning cycles \cite{sener2017active}. We take a moment to remind the reader that active learning is well suited for large datasets ($>100,000$ images) and we use these experiments to draw intuition to support the transferability of active learning for human pose estimation to practical applications.

\par
\textbf{Dataset}: The MPII \cite{mpii} dataset contains images capturing every day human activity. In contrast, the LSP-LSPET \cite{lspet} dataset represents sports poses, such as those encountered in athletics and parkour. We use standard testing splits: Newell validation split \cite{hg}, \cite{learnloss} for the MPII dataset, and the latter thousand images of the LSP dataset \cite{lspet} for the combined LSP-LSPET dataset. While the MPII dataset serves to measure the general performance of our algorithms, the LSP-LSPET dataset truly represents active learning as a process in industrial applications. Real world applications imply non-stationarity of data, hence the training  as well as validation set at any time instant do not represent the entire dataset (which is not the case with the Newell validation split). The LSP-LSPET experiment allows us to study this characteristic, where our initial training and testing pool consists of samples drawn only from the LSP dataset, with subsequent stages allowing the model to sample from the similar yet different LSPET data distribution. This represents a challenge to active learning algorithms to generalize to changes in the data distribution based on the data which is already annotated. 
\par

\begin{table*}
\large
\centering
\resizebox{2.05\columnwidth}{!}{%
\begin{tabular}{lllllp{3cm}|lllll}

\, &  \multicolumn{10}{l}{\Large \hspace*{-15mm} (a) Failure Detection: PCK scores for the images sampled at Stage $n$. (Lower PCK values indicate better identification of faulty inferences.)}\\
\toprule
\, &  \multicolumn{5}{c|}{LSP-LSPET (PCK@0.2)} & \multicolumn{5}{c}{MPII (PCKh@0.5)}\\
\midrule
\# images & 2000 & 3000 & 4000 & 5000 & 6000 & 1000 & 2000 & 3000 & 4000 & 5000\\ 
\midrule
Random & 0.430 \rpm 0.017 & 0.527 \rpm 0.012 & 0.593 \rpm 0.007 & 0.624 \rpm 0.009 & 0.645 \rpm 0.007 & 0.663 \rpm 0.012 & 0.739 \rpm 0.013 & 0.766 \rpm 0.003 & 0.792 \rpm 0.007 & 0.797 \rpm 0.006\\

Coreset & 0.288 \rpm 0.017 & 0.438 \rpm 0.020 & 0.447 \rpm 0.017 & \textbf{0.493} \rpm 0.013 & \textbf{0.556} \rpm 0.010 & 0.384 \rpm 0.014 & 0.522 \rpm 0.009 & 0.608 \rpm 0.012 & 0.697 \rpm 0.009 & 0.755 \rpm 0.029\\

LL & 0.305 \rpm 0.013 & 0.253 \rpm 0.021 & \textbf{0.358} \rpm 0.025 & 0.520 \rpm 0.011 & 0.617 \rpm 0.017 & 0.311 \rpm 0.036 & 0.465 \rpm 0.024 & 0.621 \rpm 0.017 & 0.735 \rpm 0.012 & 0.777 \rpm 0.010\\
\midrule
LL++ & \textbf{0.250} \rpm 0.011 & \textbf{0.186} \rpm 0.022 & 0.385 \rpm 0.011 & 0.533 \rpm 0.020 & 0.627 \rpm 0.012 & \textbf{0.291} \rpm 0.022 & \textbf{0.439} \rpm 0.018 & \textbf{0.610} \rpm 0.020 & \textbf{0.705} \rpm 0.023 & \textbf{0.762} \rpm 0.014\\

LL++conv & \textbf{0.209} \rpm 0.018 & \textbf{0.214} \rpm 0.028 & 0.400 \rpm 0.010 & 0.545 \rpm 0.011 & 0.635 \rpm 0.012 & \textbf{0.309} \rpm 0.029 & \textbf{0.439} \rpm 0.011 & \textbf{0.603} \rpm 0.016 & \textbf{0.704} \rpm 0.022 & 0.777 \rpm 0.008\\
\bottomrule
\, &  \multicolumn{10}{c}{\Large \,}\\

\, &  \multicolumn{10}{c}{\Large \hspace*{-20mm} (b) Testing performance: PCK scores for the testing dataset after each sampling iteration. (Higher values are better)} \\
\toprule
\, &  \multicolumn{5}{c|}{LSP-LSPET (PCK@0.2)} & \multicolumn{5}{c}{MPII (PCKh@0.5)}\\
\midrule
\# images & Stage 1 & Stage 2 & Stage 3 & Stage 4 & Stage 5 & Stage 1 & Stage 2 & Stage 3 & Stage 4 & Stage 5\\ 
\midrule
Random & 0.803 \rpm 0.003 & 0.818 \rpm 0.002 & 0.827 \rpm 0.003 & 0.834 \rpm 0.003 & 0.841 \rpm 0.002 & 0.76 \rpm 0.006 & 0.783 \rpm 0.006 & 0.803 \rpm 0.009 & 0.813 \rpm 0.004 & 0.822 \rpm 0.007\\

Coreset & 0.797 \rpm 0.008 & 0.814 \rpm 0.004 & 0.822 \rpm 0.004 & 0.831 \rpm 0.004 & 0.837 \rpm 0.003 & 0.766 \rpm 0.006 & 0.792 \rpm 0.007 & 0.812 \rpm 0.007 & 0.822 \rpm 0.011 & 0.83 \rpm 0.011\\

LL & 0.796 \rpm 0.004 & 0.814 \rpm 0.003 & 0.823 \rpm 0.004 & 0.833 \rpm 0.003 & 0.842 \rpm 0.005 & 0.763 \rpm 0.007 & 0.793 \rpm 0.005 & 0.814 \rpm 0.004 & 0.829 \rpm 0.004 & 0.838 \rpm 0.005\\
\midrule
LL++ & 0.798 \rpm 0.005 & 0.810 \rpm 0.003 & 0.824 \rpm 0.006 & 0.831 \rpm 0.004 & 0.841 \rpm 0.002 & 0.763 \rpm 0.007 & 0.791 \rpm 0.008 & 0.811 \rpm 0.005 & 0.822 \rpm 0.005 & 0.827 \rpm 0.009\\

LL++conv & 0.798 \rpm 0.005 & 0.813 \rpm 0.002 & 0.825 \rpm 0.003 & 0.832 \rpm 0.005 & 0.840 \rpm 0.005 & 0.760 \rpm 0.009 & 0.790 \rpm 0.008 & 0.808 \rpm 0.005 & 0.828 \rpm 0.003 & 0.836 \rpm 0.006\\
\bottomrule
%
&  \multicolumn{10}{c}{\Large } \\
%
%
\end{tabular}
}
\caption{LearningLoss++ outperforms others methods in failure identification Tab:(a), with similar testing accuracy as others Tab:(b). Identifying and fixing faulty inferences (high predicted loss) improves reliability of the model in open world use cases. Detecting poor inferences also allows for specific use case based collection of data, removing the need for large scale data collection in an attempt to identify model failure cases.}
\label{tab:res}
\vspace*{-1.5mm}
\end{table*}

\textbf{Experiment Design}: We simulate active learning cycles using the stacked hourglass model by training an initial base model on 1000 images, with each active learning stage selecting a new set of 1000 images from the remaining unlabelled pool of training data. The initial 1000 images for MPII are randomly sampled from the training data. whereas the initial 1000 images for LSP-LSPET are the first 1000 images from the LSP dataset (LSP consists of 2000 images, split 1000 for train and remaining form the testing set). We follow standard evaluation metrics by using PCKh@0.5 to evaluate model performance on MPII and PCK@0.2 for the LSP-LSPET images. We repeat each experiment five times and report the mean and standard deviation for the same. We limit our experiments to single person pose estimation, hence extract multiple persons in an image into separate images using the ground truth / location information from the datasets.
\par
\textbf{Failure identification}: A key question remains: \textit{How do we quantify model failures}? To eliminate subjective bias when qualitatively defining poor performance, we use the PCK/PCKh accuracy metric to identify poor performance \emph{over a set of images}. A low PCK/PCKh score corresponds to the model consistently drawing wrong inferences for that particular set of images. Our justification for using PCK/PCKh is that since these are universally accepted metrics to quantify the performance of the model, they can also be used to identify the degree to which the models perform poorly on the learning loss sampled images.

\textbf{Comparison Algorithms}: We compare our work with Learning Loss and Coreset. As noted in the related work section, DeepPrior by Caramau \etal \cite{handpose} uses dropouts as well as directly regress the joint coordinates, both of which prevent its application in human pose estimation. Entropy based approaches \cite{liu2017active} have been compared with Learning Loss \cite{learnloss} where the latter performs better than the former in identifying images with high true losses.

\textbf{Results}: We first compare the correlation between the predicted losses and the true losses for Learning Loss and LearningLoss++, which is shown in Fig: \ref{fig:corr}. The graph is computed for all the images from the LSPET dataset, with the model trained on the initial pool of images from the LSP dataset. Since the model is trained on LSP dataset and not LSPET, \textit{both the approaches are tested on their ability to identify failures and generalize to datasets from a different distribution}. We observe that the true loss mean of the images sampled with LearningLoss++ is higher as well as exhibits a lower variance among the true loss values in comparison to Learning Loss, implying that LearningLoss++ consistently identifies images where the model performs poorly. While Learning Loss performs better than entropy based approaches (shown in \cite{learnloss}), our method improves upon Learning Loss to better identify images with a high value of true loss. Similar behaviour, albeit with a smaller performance gap is obtained over the MPII dataset.
\par
The results of the simulations for the active learning cycles are shown in Table: \ref{tab:res}. We identify two phases when training a deep learning model: development and saturation phase. The development phase forms the first few cycles where the model has not converged with respect to the validation dataset. In comparison, the saturation stage marks the convergence of the model on the validation dataset, as the training dataset effectively represents the validation dataset. In practice, our models are rarely saturated because of the shift in distribution of the new incoming data as well as the potentially significant number of failure cases that arise during open world usage. Therefore, the development phase dominates model training and is of practical interest to us. Table: \ref{tab:res}(a) highlights the performance of various methods in failure detection. We see that both LearningLoss++ variants return lower PCK scores during the first few stages for both the datasets, indicating successful identification of images where the inference is faulty. With the larger MPII dataset, this trend continues into the model saturation phase. However, the smaller LSPET dataset has a limited number of \textit{tough} examples which are identified early on by LearningLoss++ as well as the original Learning Loss module. The lack of other difficult images cause a sharp rise in the PCK scores for the subsequent stages. We quickly note that detecting failures accurately does not necessary improve general model performance; failure cases are usually sparsely represented in the testing dataset which is why a quantifiable increase in accuracy is not detected. We also focus on the results in Stage 1, since stage-1 sampling is performed on the base models shared by all approaches. The ability of the convolutional network to generalize to LSPET images is highlighted in the Stage 1 results for LSP-LSPET dataset. With a PCK@0.2 value of 0.209, the convolutional network (LL++conv) comfortably outperforms its non convolutional counterpart (LL++) which has a PCK@0.2 value of 0.250. We attribute the superior performance of the convolutional architecture in LSP-LSPET to the complex poses that are usually encountered in sporting events. These complex poses are efficiently modeled by the convolutional architecture since it maintains spatial dependencies at multiple scales.

\section{Conclusion}
This work provides a strong mathematical foundation for Learning Loss, a popular active learning technique. We develop insights into the training of the learning loss module, and propose LearningLoss++ that uses an equivalent KL divergence based objective with additional benefits. Not only is the proposed objective free of any hyperparameters, the resultant gradient is smooth, which allows for better detection of faulty inferences. We also propose a convolutional architecture that exploits the spatial dependency in human pose estimation model. Our experiments show that LearningLoss++ delivers a strong performance in continuous model refinement through identification and early detection of faulty model inferences. \\
\par
\noindent
\textbf{Acknowledgment}: We thank Brijesh Pillai and Partha Bhattacharya at Mercedes-Benz R\&D India for providing the funding and compute hardware required for this work.

\newpage

{\small
\bibliographystyle{ieee_fullname}
\bibliography{egbib}
}

\end{document}


\onecolumn
\pagenumbering{gobble}

\title{Supplementary Material \\ A Mathematical Analysis of Learning Loss for Active Learning in Regression}

\author{Megh Shukla \\
Mercedes-Benz Research and Development India\\
{\tt\small megh.shukla@daimler.com}
}

\maketitle

\begin{abstract}
   This supplementary material contains the derivations that support the content of the main paper. We first derive a known result that under certain conditions the integral of the gamma distribution has a closed form solution. This solution is useful in computing the probability of sampling a pair of values within $\delta$ (Eq: 5, main paper). We then derive the gradient (Eq: 4, main paper) and its expectation (Eq: 6, main paper) of the LearningLoss++ objective.
\end{abstract}
\section{ \textbf{Integral of the Gamma Distribution} }
\vspace*{5mm}
Our goal is to compute the integral of the gamma distribution:
\begin{equation}
    \int \gamma(x, k, \Theta) \mathrm{d}x = - \Theta \sum_{n=1}^{k} \dfrac{x^{n-1}e^{- \frac{x}{\Theta}}}{\Theta^n \Gamma(n)} = G(x, k, \Theta)
    \label{eq:gamma_int}
\end{equation}
Although this is a known result, we provide a brief outline of the proof. The assumption is that $k$, the shape parameter $\in \mathbb{Z^{+}}$. We need to simplify the following:

\begin{equation}
    \int \gamma(x, k, \Theta) \mathrm{d}x = \int \dfrac{x^{k-1}e^{- \frac{x}{\Theta}}}{\Theta^k \Gamma(k)}\mathrm{d}x
    \nonumber
\end{equation}
Using integration by parts, 
\begin{equation}
    \int \gamma(x, k, \Theta) \mathrm{d}x = - \Theta  \dfrac{x^{k-1}e^{- \frac{x}{\Theta}}}{\Theta^k \Gamma(k)} + \int \dfrac{x^{k-2}e^{- \frac{x}{\Theta}}}{\Theta^{k-1} \Gamma(k-1)}\mathrm{d}x
    \nonumber
\end{equation}
The above equation can be written as $\int \gamma(x, k, \Theta) \mathrm{d}x = - \Theta \gamma(x, k, \Theta) + \int \gamma(x, k-1, \Theta) \mathrm{d}x$. By recursively solving the integral term using integration by parts, the equation reduces to $\int \gamma(x, k, \Theta) \mathrm{d}x = - \Theta \gamma(x, k, \Theta) - \Theta \gamma(x, k-1, \Theta) \ldots - \Theta \gamma(x, k=1, \Theta)$. Hence, we can write the final form of the equation as: 
\begin{equation}
    \int \gamma(x, k, \Theta) \mathrm{d}x = - \Theta \sum_{n=1}^{k} \dfrac{x^{n-1}e^{- \frac{x}{\Theta}}}{\Theta^n \Gamma(n)}
    \label{eq:gamma_int_final}
\end{equation}
\newpage
\section{\textbf{Closed form solution for $P(|X-Y| \le \delta)$ when $X, Y \sim \gamma(k, \Theta)$, $k \in \mathbb{Z}^{+}$}}
\vspace*{5mm}
The probability of sampling two variables within a margin $\delta$ of each other can be written as:
\begin{equation}
        P(|X - Y| \le \delta) = \int_0^{\delta}\gamma(x, k, \Theta)\int_0^{x+\delta}\gamma(y, k, \Theta)\mathrm{d}y\mathrm{d}x + \int_{\delta}^{\infty}\gamma(x, k, \Theta)\int_{x-\delta}^{x+\delta}\gamma(y, k, \Theta)\mathrm{d}y\mathrm{d}x 
\end{equation}
Using the previous result Eq: \ref{eq:gamma_int_final}, we can simplify the above equation as:

\begin{align}
        & = \int_0^{\delta}\gamma(x, k, \Theta) \left[ - \Theta \sum_{n=1}^{k} \dfrac{(x+\delta)^{n-1}e^{- \frac{(x+\delta)}{\Theta}}}{\Theta^n \Gamma(n)}\right] \mathrm{d}x + \int_0^{\delta}\gamma(x, k, \Theta)\mathrm{d}x \nonumber \\
        & + \int_{\delta}^{\infty}\gamma(x, k, \Theta) \left[ - \Theta \sum_{n=1}^{k} \dfrac{(x+\delta)^{n-1}e^{- \frac{(x+\delta)}{\Theta}}}{\Theta^n \Gamma(n)} \right] \mathrm{d}x - \int_{\delta}^{\infty}\gamma(x, k, \Theta) \left[ - \Theta \sum_{n=1}^{k} \dfrac{(x-\delta)^{n-1}e^{- \frac{(x-\delta)}{\Theta}}}{\Theta^n \Gamma(n)} \right] \mathrm{d}x \nonumber \\
    \label{eq:xy_le_del2}
\end{align}%

We write Eq: \ref{eq:xy_le_del2} as $P(|X - Y| \le \delta) = A + B + C + D$. While a closed form solution can be easily obtained for B (integral of gamma), We rely on the use of the binomial theorem  $(x + \delta)^{n-1} = \sum_{i=0}^{i=n-1} \Mycomb[n-1]{i} x^i \delta^{(n-1)-i}$ to simplify terms A, C and D. The method to solve A, C and D remains the same, hence we show here to focus on reducing A to a closed form solution in this supplementary material.

\begin{align}
    A & = \int_0^{\delta}\gamma(x, k, \Theta) \left[ - \Theta \sum_{n=1}^{k} \dfrac{(x+\delta)^{n-1}e^{- \frac{(x+\delta)}{\Theta}}}{\Theta^n \Gamma(n)}\right] \mathrm{d}x \nonumber \\
    & = - \Theta \sum_{n=1}^{k} \sum_{i=0}^{n-1} \int_{x=0}^{\delta} \dfrac{ \Mycomb[n-1]{i} \delta^{(n-1)-i} x^{k+i-1} e^{-\frac{2x+\delta}{\Theta}}}{\Theta^{k+n} \Gamma(n)\Gamma(k)}\mathrm{d}x
    \nonumber
\end{align}

We try to write the above equations by creating a new gamma distribution. After reorganizing the terms, we get:

\begin{equation}
    A = - \Theta e^{-\dfrac{\delta}{\Theta}} \sum_{n=1}^{k} \sum_{i=0}^{n-1} 
    \dfrac{\Mycomb[n-1]{i} \delta^{(n-1)-i}}{2^{k+i} \Theta^{n-i}}
    \int_{x=0}^{\delta} \dfrac{x^{k+i-1} e^{-\frac{x}{\Theta / 2}}}{(\dfrac{\Theta}{2})^{k+i} \Gamma(n)\Gamma(k)} \mathrm{d}x \nonumber   
\end{equation}

The final obstacle of writing the integral term into another gamma distribution is introducing $\Gamma(k+i)$. We use the property of gamma functions, $\Gamma(k+i) = \dfrac{(k+i-1)!}{(k-1)!}\Gamma(k)$. We reduce A to:

\begin{equation}
    A = - \Theta e^{-\dfrac{\delta}{\Theta}} \sum_{n=1}^{k} \dfrac{1}{\Gamma(n)} \sum_{i=0}^{n-1} \dfrac{\Mycomb[n-1]{i} \Myperm[k+i-1]{i} \delta^{(n-1)-i}}{2^{k+i} \Theta^{n-i}}
    \int_{x=0}^{\delta} \gamma(x, k+i, \Theta/2) \mathrm{d}x \nonumber   
\end{equation}

Fortunately, we have previously shown that a closed form solution exists to compute the integral of the gamma function. We therefore reach the final solution for A:

\begin{equation}
    A = - \Theta e^{-\dfrac{\delta}{\Theta}} \sum_{n=1}^{k} \dfrac{1}{\Gamma(n)} \sum_{i=0}^{n-1} \dfrac{\Mycomb[n-1]{i} \Myperm[k+i-1]{i} \delta^{(n-1)-i}}{2^{k+i} \Theta^{n-i}}
    (- \dfrac{\Theta}{2} \sum_{m=1}^{k+i} \dfrac{\delta^{m-1}e^{- \frac{\delta}{\Theta/2}}}{(\Theta/2)^m \Gamma(m)} + 1) \nonumber
    \label{eq:A}
\end{equation}
B, C, D involve a similar reduction process. Evaluation of terms at $lim\,\, x \rightarrow \infty$ reduces to 0 since all the terms contain $(x^n/e^x)$. We reproduce the final solution for A, B, C, D below:

\begin{align}
    A &= - \Theta e^{-\dfrac{\delta}{\Theta}} \sum_{n=1}^{k} \dfrac{1}{\Gamma(n)} \sum_{i=0}^{n-1} \dfrac{\Mycomb[n-1]{i} \Myperm[k+i-1]{i} \delta^{(n-1)-i}}{2^{k+i} \Theta^{n-i}}
    (- \dfrac{\Theta}{2} \sum_{m=1}^{k+i} \dfrac{\delta^{m-1}e^{- \frac{\delta}{\Theta/2}}}{(\Theta/2)^m \Gamma(m)} + 1) \nonumber \\
    B &= - \Theta \sum_{n=1}^{k} \dfrac{\delta^{n-1}e^{- \frac{\delta}{\Theta}}}{\Theta^n \Gamma(n)} + 1 \nonumber \\
    C &= - \Theta e^{-\dfrac{\delta}{\Theta}} \sum_{n=1}^{k} \dfrac{1}{\Gamma(n)} \sum_{i=0}^{n-1} \dfrac{\Mycomb[n-1]{i} \Myperm[k+i-1]{i} \delta^{(n-1)-i}}{2^{k+i} \Theta^{n-i}}
    (\dfrac{\Theta}{2} \sum_{m=1}^{k+i} \dfrac{\delta^{m-1}e^{- \frac{\delta}{\Theta/2}}}{(\Theta/2)^m \Gamma(m)}) \nonumber \\
    D &= \Theta e^{\dfrac{\delta}{\Theta}} \sum_{n=1}^{k} \dfrac{1}{\Gamma(n)} \sum_{i=0}^{n-1} \dfrac{\Mycomb[n-1]{i} \Myperm[k+i-1]{i} (-\delta)^{(n-1)-i}}{2^{k+i} \Theta^{n-i}}
    (\dfrac{\Theta}{2} \sum_{m=1}^{k+i} \dfrac{\delta^{m-1}e^{- \frac{\delta}{\Theta/2}}}{(\Theta/2)^m \Gamma(m)}) \nonumber \\
\end{align}

As in the paper, let $f(i, n, k, \delta, \Theta) = \dfrac{e^{-\frac{\delta}{\Theta}} \Mycomb[n-1]{i} \Myperm[k+i-1]{i} \delta^{(n-1)-i}}{2^{k+i} \Theta^{n-i}}$ and $G(x, k, \Theta) = - \Theta \sum_{n=1}^{k} \dfrac{x^{n-1}e^{- \frac{x}{\Theta}}}{\Theta^n \Gamma(n)}$. We note the C cancels the first term in A, leaving us with the final solution for $P(|X-Y| \le \delta)$ = A + B + C + D:
\begin{align}
        P(|X - Y| \le \delta) &= 1 - \Theta G(\delta, k, \Theta) - \Theta \sum_{n=1}^{n=k} \dfrac{1}{\Gamma(n)} \sum_{i=0}^{i=n-1} f(i, n, k, \delta, \Theta) \nonumber \\
        & + \Theta \sum_{n=1}^{n=k} \dfrac{1}{\Gamma(n)} \sum_{i=0}^{i=n-1} f(i, n, k, - \delta, \Theta) G(\delta, k+i, \Theta / 2)
\label{eq:finalfin}
\end{align}
\newpage
\section{ \textbf{LearningLoss++ Gradient} }
\vspace*{5mm}

We define similar notations from the paper:  $(l_i, l_j)$ represent the true loss for images $(x_i, x_j)$, the intermediate representations from the network for these images being $(\theta_i, \theta_j)$. We define the learning loss network to be $\hat{l}_i = \theta_i^Tw$ where $\hat{l}_i$ is the predicted/indicative loss for image $x_i$. We define the ground truth probability of sampling $x_i$ over $x_j$ as: $p_i = l_i / (l_i + l_j)$ and similarly for $p_j$. The network's probability of sampling $x_i$ over $x_j$ is $q_i = e^{\hat{l}_i} / (e^{\hat{l}_i} + e^{\hat{l}_j})$ with $q_j$ defined similarly. The minimizatio objective is: 

\begin{equation}
    \mathbb{L}_{loss}(w, \theta_i, \theta_j) = \textrm{KL}(p||q) = p_i \textrm{log}\dfrac{p_i}{q_i} + p_j \textrm{log}\dfrac{p_j}{q_j}
    \label{eq:plus_1}
\end{equation}

On substituting $p, q$ and computing the gradient with respect to $w$, Eq: \ref{eq:plus_1} reduces to:

\begin{align}
    \nabla_w \mathbb{L} & = - \nabla_w \left[ p_i log (\dfrac{e^{\theta_i^Tw}}{e^{\theta_i^Tw} + e^{\theta_j^Tw}}) + p_j log (\dfrac{e^{\theta_j^Tw}}{e^{\theta_i^Tw} + e^{\theta_j^Tw}}) \right] \nonumber \\
    & = - \nabla_w \left[ p_i \theta_i^Tw + p_j \theta_j^Tw - (p_i+p_j) \textrm{log}(e^{\theta_i^Tw} + e^{\theta_j^Tw}) \right] \nonumber \\
    & = - p_i\theta_i - p_j\theta_j + \dfrac{e^{\theta_i^Tw}\theta_i + e^{\theta_j^Tw}\theta_j}{e^{\theta_i^Tw} + e^{\theta_j^Tw}} \nonumber
\end{align}

Using the definition of $\hat{l}_i, \hat{l}_j, q_i, q_j$, the equation can be written as:
\begin{equation}
    \nabla_w \mathbb{L} = -p_i\theta_i - p_j\theta_j + q_i\theta_i + q_j\theta_j
\end{equation}

Since $p_i + p_j = 1$, $q_i + q_j = 1$, we get $(q_i - p_i) = -(q_j - p_j)$. 
The final gradient can now be written as:
\begin{equation}
    \nabla_w \mathbb{L}(w, \theta_i, \theta_j) = (q_i - p_i)(\theta_i - \theta_j)
    \label{eq:grad_llpp}
\end{equation}
\newpage
\section{ \textbf{Expected Gradient for LearningLoss++} }
\vspace*{5mm}
Since providing a proof for the entire solution is time consuming and lengthy, we provide a derivation for the main skeleton and show that the solutions discussed above (integral of gamma, binomial) can be reused to obtain a closed form solution for the expected gradient.
We continue from Eq: 8 in the paper; the expected gradient is defined as:

\begin{align}
    \mathbb{E}_{x}[\nabla_w \mathbb{L} (X=x, Y=x+\delta_2 \, | \, \delta_2)] = \int_{x=0}^{x= \infty}\int_{y=x+\delta_1}^{y=x+\delta_2} (q_i - \dfrac{x}{2x + \delta_2})(\theta_i - \theta_j)p(x, y | \delta_2) \mathrm{d}y \mathrm{d}x
    \label{eq:expected_1}
\end{align}

Where we define $\delta_1$ as $\textrm{lim}\,\, \delta_2 - \delta_1 \rightarrow 0^+$ to accurately define area under the curve as probability. By definition, $p(x, y | \delta_2) = \dfrac{\gamma(x,k,\Theta) \gamma(y,k,\Theta)}{p(y-x=\delta_2)}$, since $X, Y \sim \gamma(x, k, \Theta)$. Here, $p(y - x = \delta_2)$ is the normalizer.  We note that $p(y - x = \delta_2) = \int_{x=0}^{\infty} \int_{y=x+\delta_1}^{x+\delta_2} \gamma(x,k,\Theta) \gamma(y,k,\Theta) \textrm{d}y\textrm{d}x $ and $\delta_1 \rightarrow \delta_2^{-}$. We simplify $p_i = \dfrac{x}{2x + \delta_2} = \dfrac{1}{2}(1 - \dfrac{\delta_2}{2x+\delta_2})$. The expectation reduces to:

\begin{equation}
    \mathbb{E}_x[\nabla_w\mathbb{L}] = q_i(\theta_i - \theta_j) - \dfrac{(\theta_i - \theta_j)}{2} \left[ 1 - \int_{x=0}^{\infty} \dfrac{\delta_2}{2x+\delta_2} \int_{y=x+\delta_1}^{x+\delta_2} \dfrac{\gamma(x,k,\Theta) \gamma(y,k,\Theta)}{p(y-x=\delta_2)} \textrm{d}y\textrm{d}x \right]
    \label{eq:egrad1}
\end{equation}

Since $p(y - x = \delta_2)$ is the normalizer, it is constant given $\delta_2$. We therefore write $\mathbb{D} = p(y - x = \delta_2)$. This allows us to write Eq: \ref{eq:egrad1} as:

\begin{equation}
    = q_i(\theta_i - \theta_j) - \dfrac{(\theta_i - \theta_j)}{2} \left[ 1 - \int_{x=0}^{\infty} \dfrac{\delta_2}{2x+\delta_2} \dfrac{\gamma(x,k,\Theta)}{\mathbb{D}} \int_{y=x+\delta_1}^{x+\delta_2} \gamma(y,k,\Theta) \textrm{d}y\textrm{d}x \right]
    \label{eq:egrad2}
\end{equation}

We see that Eq: \ref{eq:egrad2} bears a strong resemblance with the derivation of $P(|X-Y| \le \delta)$ we proved earlier. We can directly substitute the values of $\gamma(x, k, \Theta)$, $\int_{y=x+\delta_1}^{x+\delta_2} \gamma(y,k,\Theta)$ [Integral of Gamma] and $f(i, n, k, \delta, \Theta)$ from Eq: \ref{eq:finalfin} into the above equation to get:

\begin{align}
    = q_i(\theta_i - \theta_j) - \dfrac{(\theta_i - \theta_j)}{2}\,\,\,[ \,\,\,1 &+ \dfrac{\Theta}{\mathbb{D}} \sum_{n=1}^{k} \dfrac{1}{\Gamma(n)} \sum_{i=0}^{n-1} f(i, n, k, \delta_2, \Theta) \int_{x=0}^{\infty} \dfrac{x^{k+i-1} e^{-\frac{x}{\Theta / 2}}}{(\dfrac{\Theta}{2})^{k+i} \Gamma(k+i)} \dfrac{\delta_2}{2x+\delta_2} \mathrm{d}x \nonumber \\
    &-  \dfrac{\Theta}{\mathbb{D}} \sum_{n=1}^{k} \dfrac{1}{\Gamma(n)} \sum_{i=0}^{n-1} f(i, n, k, \delta_1, \Theta) \int_{x=0}^{\infty} \dfrac{x^{k+i-1} e^{-\frac{x}{\Theta / 2}}}{(\dfrac{\Theta}{2})^{k+i} \Gamma(k+i)} \dfrac{\delta_2}{2x+\delta_2} \mathrm{d}x \,\,\,] \nonumber
\end{align}

Let $t = 2x + \delta_2$, then the above equation reduces to:

\begin{align}
    = q_i(\theta_i - \theta_j) - \dfrac{(\theta_i - \theta_j)}{2}\,\,\,[ \,\,\,1 &+ \dfrac{\Theta}{\mathbb{D}} \sum_{n=1}^{k} \dfrac{1}{\Gamma(n)} \sum_{i=0}^{n-1} f(i, n, k, \delta_2, \Theta) \int_{t= \delta_2}^{\infty} \dfrac{(t-\delta_2)^{k+i-1} e^{-\frac{(t-\delta_2)}{\Theta / 2}}}{(\dfrac{\Theta}{2})^{k+i} \Gamma(k+i)} \dfrac{\delta_2}{t} \mathrm{d}t \nonumber \\
    &-  \dfrac{\Theta}{\mathbb{D}} \sum_{n=1}^{k} \dfrac{1}{\Gamma(n)} \sum_{i=0}^{n-1} f(i, n, k, \delta_1, \Theta) \int_{t= \delta_2}^{\infty} \dfrac{(t-\delta_2)^{k+i-1} e^{-\frac{(t-\delta_2)}{\Theta / 2}}}{(\dfrac{\Theta}{2})^{k+i} \Gamma(k+i)} \dfrac{\delta_2}{t} \mathrm{d}t \,\,\,] \nonumber
\end{align}

If we let $I(k+i, \Theta) = \int_{t= \delta_2}^{\infty} \dfrac{(t-\delta_2)^{k+i-1} e^{-\frac{(t-\delta_2)}{\Theta / 2}}}{(\dfrac{\Theta}{2})^{k+i} \Gamma(k+i)} \dfrac{\delta_2}{t} \mathrm{d}t$, then we can write the expected gradient $\mathbb{E}_x[\nabla_w\mathbb{L}]$ as:

\newpage
\begin{align}
    \mathbb{E}_{x}[\nabla_w \mathbb{L} (\delta)] = (\theta_i - \theta_j) \Bigg[ q_i - \dfrac{1}{2} +  \dfrac{\Theta}{2\mathbb{D}}\sum_{n=1}^{k}\dfrac{1}{\Gamma(n)} \sum_{i=0}^{n-1}I(k+i, \Theta)[f(i, n, k, \delta_2, \Theta) - f(i, n, k, \delta_1, \Theta)] \Bigg]
    \label{eq:expected_final}
\end{align}

We note that\textbf{ this is the final expected gradient given the margin $\delta$} and $\delta_1 \rightarrow \delta_2^{-} = \delta$. However, we still need to compute the closed form solution for $\mathbb{D}$ and $I(k+i, \Theta)$.
We first compute the value of $\mathbb{D}$:

\begin{equation}
    \mathbb{D} = p(y - x = \delta_2) = \int_{x=0}^{\infty} \gamma(x,k,\Theta) \int_{y=x+\delta_1}^{x+\delta_2} \gamma(y,k,\Theta) \textrm{d}y\textrm{d}x \nonumber
\end{equation}

We have previously computed a similar result when deriving the closed form solution, where use the integral of gamma as well as the binomial theorem to solve for integrating a gamma function within a gamma function. To avoid repetitive steps, we present the final solution for $\mathbb{D}$:

\begin{equation}
    \mathbb{D} = p(y - x = \delta_2) = \Theta \sum_{n=1}^{k} \dfrac{1}{\Gamma(n)} \sum_{i=0}^{n-1} f(i, n, k, \delta_{1}, \Theta) - f(i, n, k, \delta_{2}, \Theta)
\end{equation}
The solution for $I(k+i, \Theta) = \int_{t= \delta_2}^{\infty} \dfrac{(t-\delta_2)^{k+i-1} e^{-\frac{(t-\delta_2)}{\Theta / 2}}}{(\dfrac{\Theta}{2})^{k+i} \Gamma(k+i)} \dfrac{\delta_2}{t} \mathrm{d}t$ is similar with the use of the binomial theorem to convert the integral into a sum of integrals. However, the following caveat exists: The division by $t$ renders one term in the expansion an exponential integral of the form $\dfrac{e^{-t}}{t}$. This is reflected in the solution for $I(k+i, \Theta)$:

\begin{equation}
    I(u=k+i, \Theta) = e^{\dfrac{\delta_2}{\Theta}} \sum_{j=1}^{u-1}
    \dfrac{(-1)^{u-1-j} \delta_2^{u-j}\Mycomb[u-1]{j}}{\theta^{u-j} \Myperm[u-1]{u-j}} \int_{t=\delta_2}^{\infty}\gamma(j, \Theta)\,\, +\,\, 
    \dfrac{e^{\frac{\delta_2}{\Theta}}(-1)^{u-1}\delta_2^{u}}{\Theta^u (u-1)!} \Gamma(0, \dfrac{\delta_2}{\Theta})
\end{equation}
While the first term again contains the integral of the gamma function which is a closed form solution, the second term is a consequence of the exponential integral that leads to the lower incomplete gamma function. We therefore have shown that both $\mathbb{D}$ and $I(u=k+i, \Theta)$ have closed form solutions, allowing the expected gradient Eq: \ref{eq:expected_final} to have a closed form solution.